%% file: main.tex
\documentclass{article}

\usepackage{microtype}
\usepackage{graphicx}
\usepackage{booktabs} %
\usepackage{xspace}
\usepackage{bbm}

\usepackage{hyperref}
\newif\ifcomments
\commentstrue

\newcommand{\method}{\texttt{CARE}\xspace}
\newcommand{\longmethod}{\emph{Conditional Alignment and Reweighting}\xspace}

\usepackage{floatrow}

\usepackage[accepted]{icml2023}

\usepackage{amsmath}
\usepackage{mathtools}
\usepackage{amsthm}

\usepackage{amssymb}

\usepackage{adjustbox}
\usepackage{booktabs}
\usepackage{pifont}
\usepackage{paralist}
\usepackage{multirow}
\usepackage{color, colortbl}
\usepackage{xcolor}
\usepackage{subcaption}
\usepackage[font=small,tableposition=top]{caption}
\definecolor{Gray}{gray}{0.9}

\newcommand*{\drivesim}{DriveSim\xspace}
\newcommand*{\drivesimtocityscapes}{DriveSim$\to$CityScapes\xspace}

\newcommand{\EX}{\mathbb{E}}
\newcommand{\source}{S}  %
\newcommand{\target}{T}  %
\newcommand{\model}{h}  %

\newcommand{\ie}{\emph{i.e.}~}

\usepackage{cancel}
\newcommand{\defeq}{:=}
\newcommand{\xsf}{\mathsf{x}}
\newcommand{\ysf}{\mathsf{y}}
\newcommand{\wsf}{\mathsf{w}}
\newcommand{\hsf}{\mathsf{h}}
\usepackage[capitalize,noabbrev]{cleveref}

\theoremstyle{plain}

\theoremstyle{definition}

\theoremstyle{remark}

\usepackage[textsize=tiny]{todonotes}

\icmltitlerunning{Supervised Detection Adaptation with Conditional Alignment and Reweighting}

\begin{document}

\twocolumn[
\icmltitle{Bridging the Sim2Real gap with CARE:\\Supervised Detection Adaptation with Conditional Alignment and Reweighting}

\icmlsetsymbol{equal}{*}

\begin{icmlauthorlist}
\icmlauthor{Viraj Prabhu}{sch}
\icmlauthor{David Acuna}{comp}
\icmlauthor{Andrew Liao}{comp}
\icmlauthor{Rafid Mahmood}{comp}
\icmlauthor{Marc T. Law}{comp}\\
\icmlauthor{Judy Hoffman}{sch}
\icmlauthor{Sanja Fidler}{comp}
\icmlauthor{James Lucas}{comp}
\end{icmlauthorlist}

\icmlaffiliation{comp}{NVIDIA}
\icmlaffiliation{sch}{Georgia Tech}

\icmlcorrespondingauthor{Viraj Prabhu}{virajp@gatech.edu}
\icmlcorrespondingauthor{James Lucas}{jlucas@nvidia.com}

\icmlkeywords{Machine Learning, Domain Adaptation, Computer Vision}

\vskip 0.3in
]

\printAffiliationsAndNotice{Work done while V.P. was an intern at NVIDIA.}  %

\input{sections/abstract.tex}

\input{sections/intro}

\input{sections/related}

\input{sections/method_new}

\input{sections/results}

\input{sections/conclusion}

\bibliography{main}
\bibliographystyle{icml2023}

\newpage
\appendix
\onecolumn

\input{sections/appendix}

\end{document}

%% file: sections/abstract.tex
\begin{abstract}
    Sim2Real domain adaptation (DA) research focuses on the constrained setting of adapting from a labeled synthetic source domain to an unlabeled or sparsely labeled real target domain. However, for high-stakes applications (\emph{e.g.} autonomous driving), it is common to have a modest amount of human-labeled real data in addition to plentiful auto-labeled source data (\emph{e.g.} from a driving simulator).
    We study this setting of \emph{supervised} sim2real DA applied to 2D object detection. We propose Domain Translation via Conditional Alignment and Reweighting (\method) a novel algorithm that systematically exploits target labels to explicitly close the sim2real appearance and content gaps. We present an analytical justification of our algorithm and demonstrate strong gains over competing methods on standard benchmarks.
\end{abstract}

%% file: sections/intro.tex
\section{Introduction}
\label{ref:intro}

Domain  Adaptation (DA) is a framework 
that seeks to overcome shifts in data distributions between training and testing. Typically, DA methods assume  access to a large amount of labeled data from the training (source) distribution, and unlabeled or sparingly labeled data from the test (target) distribution~\citep{saenko2010adapting,ganin2014unsupervised,tzeng2017adversarial}. DA has been extensively studied in computer vision for applications where annotating target data is expensive~\cite{csurka2022}.

\begin{figure}[!t]
  \centering 
    \includegraphics[width=\textwidth]{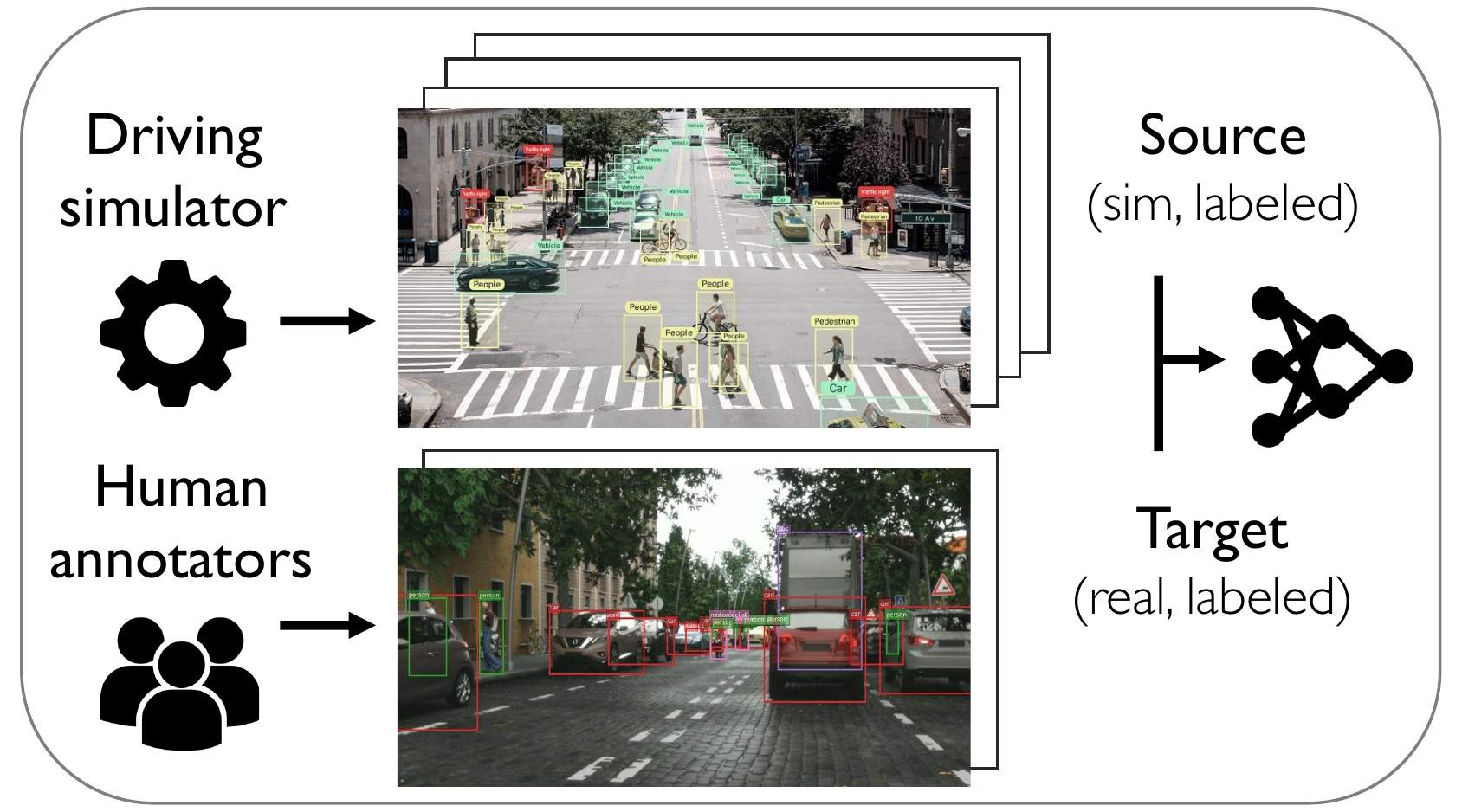}  \caption{ \label{fig:teaser}
    Traditional Sim2Real domain adaptation assumes access to very few or no target labels, which is unrealistic in high-stakes applications like self-driving. We study the practical setting of \emph{supervised} Sim2Real domain adaptation applied to 2D object detection, wherein the goal is to maximize heldout target performance given access to human-labeled target data and an additional large set of machine-labeled simulated data. 
    }
  \end{figure}

As annotation costs decrease, it becomes increasingly practical to annotate more target data, especially in high-stakes industrial applications such as autonomous driving~\citep{mahmood2022optimizing}. 
A common practice in this field is to augment a target dataset of real driving scenarios with an additional labeled dataset generated in  simulation~\cite{teslaaiday,kishore2021synthetic}. Simulated data may be particularly useful to improve performance on the long-tail of driving scenarios for which it may be difficult to challenging to collect real labeled data~\cite{rempe2022generating,resnick2022causal}.%
  In this paper, we formulate this setting as a \emph{supervised} Sim2Real DA problem. We use simulated, machine-labeled source data, and real, human-labeled target data (see Fig.~\ref{fig:teaser}), and ask: in this label-privileged setting, what would be the most effective way to combine sim and real data to improve target performance?

\begin{table}[!t]    
  \vspace{-1.5em}
  \centering  \RawFloats
  \caption{ \label{tab:teaser}
    Car detection adaptation from Sim10K$\to$Cityscapes: Systematically combining labeled source and target data improves over using a single data source as well as na\"ive combinations. 
  }
  \resizebox{\linewidth}{!}{
  \setlength{\tabcolsep}{2pt}
      \begin{tabular}{lc}
        \toprule
         \multicolumn{1}{c}{\centering \bf Method} & \multicolumn{1}{c}{\centering \bf mAP@50 ($\uparrow$)} \\
        \toprule
        Only labeled source data & 41.8 \\
        Only labeled target data & 62.1 \\     
        UDA~\cite{khindkar2022miss} (Labeled source $+$ unlabeled target) & 53.1 \\
        FDA~\cite{wang2020frustratingly} (Labeled source $+$ labeled target) & 65.2 \\
        Mixing~\cite{kishore2021synthetic} (Labeled Source $+$ labeled target) & 64.8 \\
        Seq. FT~\cite{tremblay2018training} (Labeled source $+$ labeled target) & 66.4 \\        
        \rowcolor{Gray}
        Ours - \method (Labeled source $+$ labeled target) & \textbf{68.1} \\
        \bottomrule
        \end{tabular}}
  \vspace{-1.5em}
\end{table}

Surprisingly, this practical setting has received little interest in recent domain adaptation literature, which focuses on  unsupervised adaptation~\emph{(no target labels,}~\citet{chen2018domain,acuna2021towards,li2022cross}), and few-shot and semi-supervised adaptation (\emph{few target labels},~\citet{donahue2013semi,wang2019few,saito2019semi,wang2020frustratingly}).
Although such methods could be extended to the supervised setting, \emph{e.g.} by adding a supervised target loss to an off-the-shelf unsupervised DA method, 
we find this to be suboptimal in practice (see Table \ref{tab:teaser}), since these straightforward extensions do not exploit large-scale target labels and their statistics for domain alignment.
Similarly, few-shot and semi-supervised adaptation methods assume access to limited target labels (\emph{e.g.} 8 labeled images per class for object detection, \citet{wang2019few}) that are insufficient for reliably estimating target statistics. 
 Facing this research gap, industry practitioners may resort to na\"ively combining labeled source and target data via mixing~\cite{kishore2021synthetic} (\emph{i.e.} training on combined source and target data) or sequential fine-tuning~\cite{tremblay2018training,prakash2019structured,prakash2021self} (\emph{i.e.} training on source data followed by finetuning on target data). However, these simple heuristics do not address the domain gap between simulation and reality.

This paper addresses the research-practice gap to show that \emph{systematically} combining the two labeled data sets can significantly improve performance over competing methods (see Table \ref{tab:teaser}). 
We propose a general framework called \emph{Domain Translation via Conditional Alignment and Reweighting} (\method) for supervised Sim2Real DA. 
\method builds on commonly-used baselines and off-the-shelf adaptation methods but explicitly leverages existing labels in the target domain to minimize both appearance (pixel and instance-level visual disparity) and content gaps (disparities in task label distributions and scene layout). Specifically, we overcome the appearance gap by explicitly using ground-truth labels to conditionally align intermediate instance representations. To overcome the content gap, we conditionally reweight the importance of samples using estimated spatial, size, and  categorical distributions. We formalize  our setting using the joint risk minimization framework, and provide theoretical insights for our  design choices. Finally, we apply our framework to the challenging task of 2D object detection 
We make the following contributions:

(1) We present a detailed study of supervised Sim2Real object detection adaptation and show that existing methods yield suboptimal performance by not adequately exploiting target labels. (2) We propose \method, a general framework for supervised Sim2Real domain adaptation and apply it to the 2D object detection. On three standard Sim2Real benchmarks for detection adaptation, \method strongly outperforms competing methods (\emph{e.g.} boosting mAP@50 by as much as $\sim$25\% on Synscapes$\to$Cityscapes). (3) We formalize our setting using the joint risk minimization framework and provide theoretical insights into our design choices.

%% file: sections/related.tex
\section{Related work}
\label{ref:related}

To our knowledge, supervised domain adaptation (SDA) for object detection has not seen recent work in computer vision. Early DA works~\citep{saenko2010adapting,kulis2011you,hoffman2013efficient,tsai2016learning} have studied the SDA setting applied to image classification, proposing contrastive-style approaches based on metric learning with cross-domain pairwise constraints. However, these works predate deep learning and do not study complex tasks like object detection. Below, we summarize lines of work in the related areas of unsupervised and few-shot adaptation.

\textbf{Unsupervised domain adaptation (UDA)}. 
The DA literature primarily focuses on \emph{unsupervised} adaptation from a labeled source setting to an unlabeled target domain~\cite{saenko2010adapting,ganin2014unsupervised,hoffman2018cycada}. Successful UDA approaches have employed different strategies ranging from domain adversarial learning~\cite{long2015learning,pmlr-v139-acuna21a} to domain discrepancy minimization~\cite{long2018conditional}, image translation~\cite{hoffman2018cycada}, and self-training~\cite{prabhu2021sentry,li2022cross}. Cross-domain object detection has also seen recent work, based on multi-level domain adversarial learning~\citep{chen2018domain}, strong-weak distribution alignment of local and global features~\citep{saito2019strong}, and domain adversarial learning weighted by region discriminativeness \citep{zhu2019adapting}, Alternatively, \citet{roychowdhury2019automatic,li2022cross} self-train with refined pseudolabels, and \citet{kim2019self} use background regularization.

Importantly, due to the absence of target labels, UDA methods resort to approximations based on marginal alignment or pseudolabels. In this paper, we instead consider \emph{supervised} Sim2Real adaptation where ground-truth labels are provided for the target dataset during training. To compare against our approach, we benchmark supervised extensions of existing UDA methods as baselines in our paper.%

\begin{figure*}[t]
    \centering
    \includegraphics[width=1.0\textwidth]{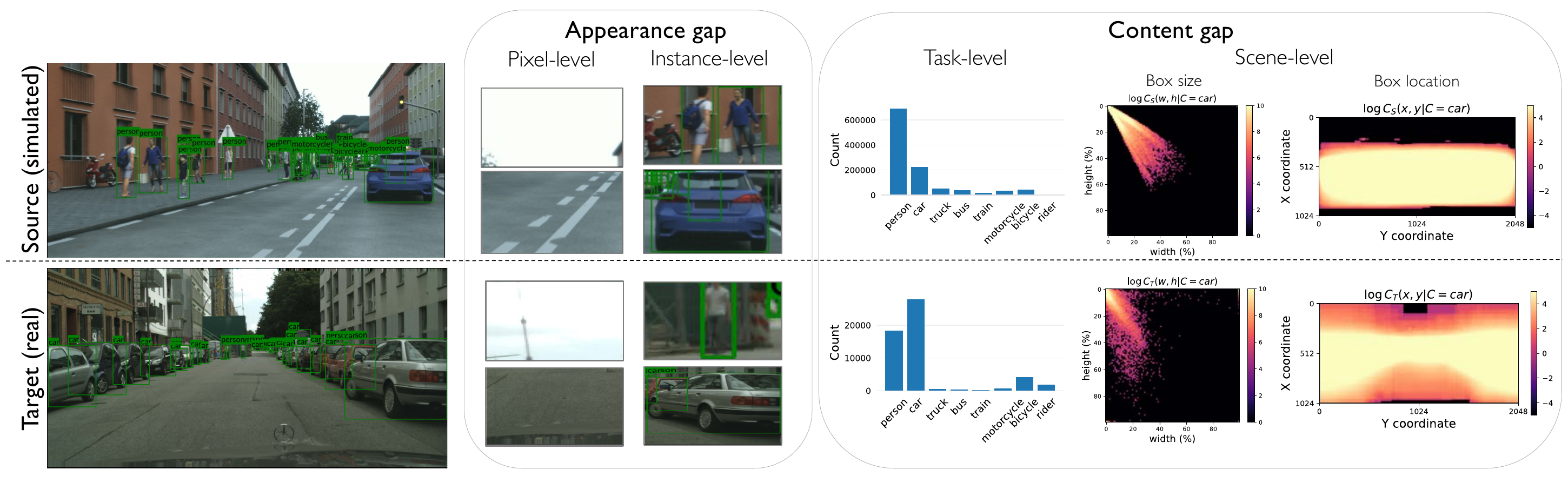}
    \caption{ \label{fig:domain_gap}
      The domain gap between a simulated source and real target domain consists of an appearance and content gap. The appearance gap corresponds to pixel-level differences (\emph{e.g.} texture and lighting) and instance-level differences (\emph{e.g.} vehicle design). The content gap consists of differences in label distributions due to different class frequencies and bounding box sizes and locations. \textbf{Right.} Column 1: Task label histograms. Column 2: Empirical distribution of ``car'' box \emph{sizes}. Column 3: Empirical distribution of ``car'' box \emph{locations}.
    }
    
  \end{figure*}

\textbf{Few-shot (FDA) and Semi-supervised Domain Adaptation (SSDA).} 
Closer to our setting are Few-shot DA learning (FDA,~\citet{wang2019few,gao2022acrofod,zhong2022pica,ramamonjison2021simrod}) and Semi-supervised DA(SSDA,~\citet{donahue2013semi,yao2015semi,saito2019semi}), which differ in important ways.
FDA assumes a very small amount of labeled target data is available (\emph{e.g.} 8 images per-class for detection in~\citet{wang2019few}). Such methods employ source feature-regularized images with instance-level adversarial learning~\cite{wang2019few}, point-wise distribution alignment~\cite{zhong2022pica}, and multi-level domain-aware data augmentation~\cite{gao2022acrofod}.
SSDA also assumes limited target labels (\emph{e.g.} 1 to 3 images per category for image classification~\cite{saito2019semi}), but additionally leverages a large set of \emph{unlabeled} target data, making use of min-max entropy optimization~\cite{saito2019semi} or student-teacher learning frameworks~\cite{li2022cross}. 
Instead, we operate in a \emph{supervised} DA setting with access to a substantial amount of labeled target data in addition to a large (in theory, possibly infinite) amount of labeled simulated data. As a result, SDA uniquely permits \emph{reliable} estimates of target statistics. Our algorithm leverages these statistics and target labels to systematically close the Sim2Real domain gap.

%% file: sections/method_new.tex
\section{Approach}
\label{sec:method}

In this section, we first introduce the supervised Sim2Real detection adaptation problem (Section \ref{subsec:formulation}). 
We characterize two primary aspects of the Sim2Real domain gap: an appearance and a content gap (Section \ref{subsec:domain_gap}).
Finally we introduce our method \method that leverages a labeled target dataset to close this domain gap (Section \ref{subsec:closing_domain_gap}) and provide an analytical justification of the algorithm (Section \ref{subsec:theoretical_interpretation}).

\subsection{Problem Formulation}
\label{subsec:formulation}

Let $\mathcal{X}$ and $\mathcal{Y}$ denote input and output spaces. 
In object detection, $x \in \mathcal{X}$ are images ($\mathcal{X}  \subseteq \mathbb{R}^{H\times W \times 3}$) and $y := (B, C) \in \mathcal{Y}$ are $K$-class labels with $C \in \{1, .., K\}$ and bounding boxes $B \subseteq \{ (\wsf, \hsf, \xsf, \ysf) \in \mathbb{R}^{4} \}$ (comprising the width $\wsf$, height $\hsf$, and centre coordinates $(\xsf, \ysf)$, respectively). 
Let $\model(x) \defeq h_\theta(g_\phi(x))$ be an object detector composed of a feature extractor $g(x)$ and a classifier $h(g(x))$ that are parameterized by $\phi$ and $\theta$.
Matching prior object detection work \citep{khindkar2022miss, wang2021robust}, we design $h(g(x))$ via Faster RCNN \citep{ren2015faster}, which uses a region proposal network that receives features generated by a backbone network and passes them through an ROI align layer to obtain ROI features; these are then passed through a final box predictor.
We let $\hat{B}, \hat{C} = \arg\max h(g(x))$ be bounding box coordinates and object class predicted by the model for input image $x$.
In sim2real SDA, we are given two labeled data sets representing a (simulated) source distribution $P_\source$ and a (real) target distribution $P_\target$. %
Our goal is to minimize the expected risk of a detection loss consisting of a classification loss $\ell_{cls}$ and bounding box regression loss $\ell_{box}$:
\begin{equation} \label{eq:l_det}
\ell_{det}(h(g(x)), B, C) \defeq \ell_{box}(\hat{B}, B) + \ell_{cls}(\hat{C}, C)
\end{equation}
over a target domain $r_\target \defeq \EX_{x, B, C \sim P_\target} [ \ell_{det}(h(x), B, C) ]$.

\subsection{Characterizing the Sim2Real Domain Gap}
\label{subsec:domain_gap}

Leveraging the source distribution to improve performance on the target is challenging due to the \emph{domain gap} which exists in both the image and label distributions. 
We partition this gap into two categories: appearance and content gap~\cite{kar2019meta} and characterize these in detail, using the Synscapes~\cite{wrenninge2018synscapes}$\to$ Cityscapes~\cite{cordts2016cityscapes} shift for object detection adaptation as an example. 

The \textbf{appearance gap} consists of visual disparities between images from the two domains (see Fig \ref{fig:domain_gap}, \emph{left}). For example, a pixel-level appearance gap may be due to differences in lighting between real and simulated images~\cite{chattopadhyay2022pasta}, while an instance-level gap may be due to differences in the appearance of synthesized versus real objects. 
We characterize the appearance gap as the dissimilarity $D(\cdot,\cdot)$ in the probabilities between source and target distributions when conditioned on the label (e.g. $ D(P_\source(x|B,C), P_\target(x|B,C)) $). 

The \textbf{content gap} can be decomposed into scene-level changes in the layout of objects (e.g., size and spatial distribution) as well as shifts in the task label distributions and the frequencies of classes (see Fig \ref{fig:domain_gap}, \emph{right}). 
We characterize the scene-level changes as the dissimilarity in the probabilities of object bounding boxes when conditioned on the class $D(P_\source(B|C), P_\target(B|C))$ and the task-level class frequency gap as the dissimilarity in class probabilities $D(P_\source(C), P_\target(C))$.

\begin{figure}[t] \label{fig:approach}
    \centering
    \includegraphics[width=1.0\textwidth]{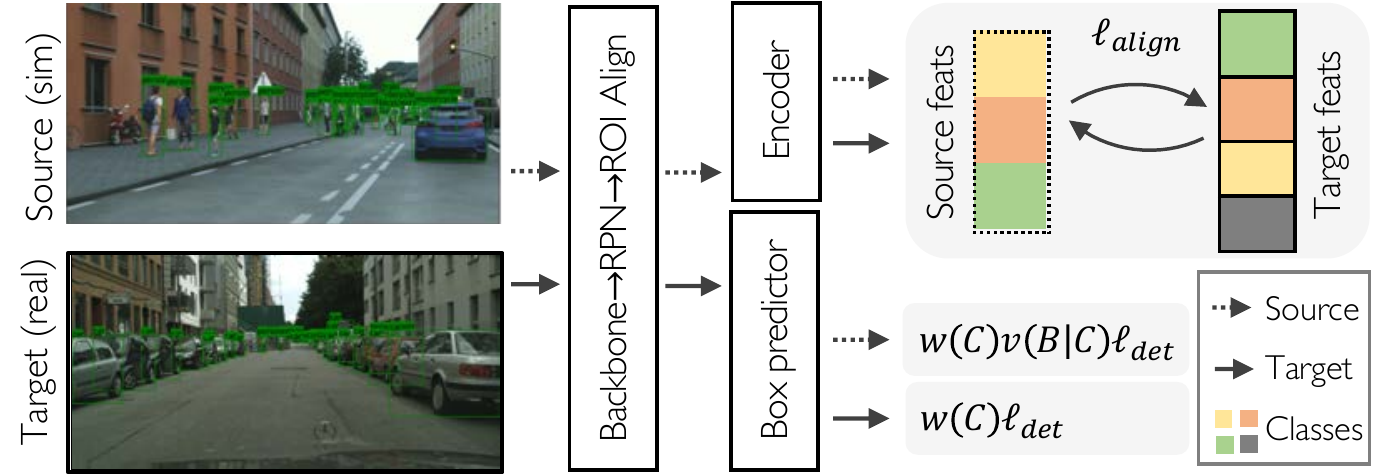}
    \caption{
      Conditional Alignment and Reweighting (\method) exploits target labels to estimate and bridge cross-domain appearance gaps (via a cycle consistency-based conditional feature alignment objective) and content gaps (via importance reweighting).
    }
  \end{figure}

\subsection{Bridging the domain gap with \method} \label{subsec:closing_domain_gap}

To close the sim2real gap, \longmethod  (\method) 
minimizes the effect of both the appearance and the content gap via feature alignment and importance reweighing. 
Let $w_S(C) \defeq 1 / P_S(C), w_T(C) \defeq 1 / P_T(C)$ be the inverse class frequency for each domain and let $v(B|C) \defeq P_\target(B|C)/P_\source(B|C)$ be the inverse ratio of the scene-level bounding box frequency gap. 
These reweighting factors ensure that the learned classifier considers that the source and target data sets follow the same distribution during training. 
In \method, we minimize the following \emph{domain translation} loss:
\begin{align}
\label{eq:dsda}
    \begin{split}
     \min_{\theta, \phi}&\ \EX_{x, B, C \sim P_\source} \bigg[ w_\source(C) v(B|C) \ell_{det}(h(g(x)), B, C)\bigg] \\ 
& + \EX_{x', B', C' \sim P_\target} \big[ w_T(C') \ell_{det}(h(g(x')), B', C') \big] \\ 
& + \lambda \EX_{\genfrac{}{}{0pt}{}{x', B', C' \sim P_\target}{x , B, C \sim P_\source}} \bigg[ \ell_{align}(g(x), g(x')) \bigg| C=C' \bigg].
    \end{split}
\end{align}
where $\ell_{align}$ is defined in Eq. \eqref{eq:l_align}, and $\lambda \geq 0$ is a regularization parameter. 
The above loss minimizes three terms, where the first term is a reweighted detection loss over the source dataset and the second loss is a class-balanced detection loss over the target dataset. 
The third term aligns the encoded features $g(x)$ and $g(x')$ of similar cross-domain instance embeddings belonging the same class. We now elaborate upon each term.

\subsubsection{Bridging appearance gap with cross-domain cycle consistency}

\begin{figure}[!t]
  \centering
  \includegraphics[width=0.9\textwidth]{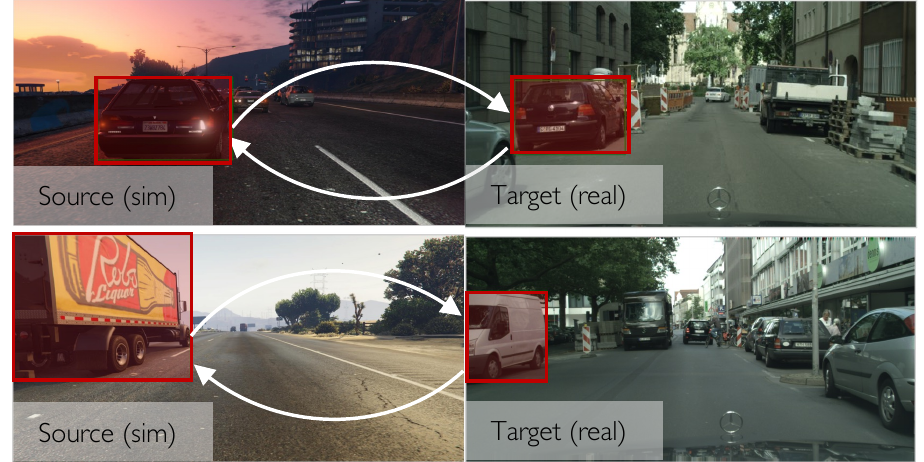}
  \caption{
    Visualization of cross-domain cycle consistency matching with \method on Sim10K$\to$Cityscapes. \method embeds similar-looking cars closer to minimize the appearance gap.
  }
  \label{fig:viz}
\end{figure}

To minimize the appearance gap, $\ell_{align}$ performs a class-and-box conditional feature alignment strategy by optimizing a cross-domain cycle consistency objective.
Specifically, we extract ROI features corresponding to the ground truth bounding box coordinates of both source and target images and match similar cross-domain instance features belonging to the same class.
Fig. \ref{fig:viz} visualizes the intuition.

For a given class, suppose we are given $k$ ground truth bounding boxes from the source and target domains each. For each instance, our encoder extracts $d$-dimensional ROI features $\mathbf{f}^i_\omega \in \mathbb{R}^d$, where $i \in \{1,\dots,k\}$ and $\omega \in \{\source, \target\}$ denote the $i$-th feature and the domain, respectively. 
We first measure the (negative of the) squared Euclidean distance between these same-class cross-domain features: 
\begin{equation}\nonumber
    s_{i, j}\defeq - \|\mathbf{f}_S^i-\mathbf{f}_T^j \|_2^2.
\end{equation}
For each target $j$, we compute soft-matching features
\begin{equation}\nonumber
\hat{\mathbf{f}}_T^j\defeq\sum_{j'=1}^{k} \alpha_{j, j'} \mathbf{f}_T^{j'}, ~\text{where }~ \alpha_{j, j'}\defeq\frac{e^{s_{j, j'}}}{\sum_{m=1}^{k} e^{s_{j, m}}}
\end{equation}
Finally, we assemble a similarity score between each source $i$ and target $j$ instance by minimizing the negative squared Euclidean distance between the source and the soft-matching target feature vectors 
\begin{equation}\nonumber
\hat{s}_{i, j}\defeq - \| \mathbf{f}_S^i-\hat{\mathbf{f}}_T^j \|_2^2.
\end{equation}
Let $\mathbf{\hat{s}}^j \defeq \left[\hat{s}_{1, j}, \dots, \hat{s}_{k, j}\right]$ be the vector of similarity scores for the $j$-th target. 
Our cycle matching alignment loss minimizes the cross entropy between features as follows:
\begin{align} \label{eq:l_align}
    \resizebox{0.91\hsize}{!}{
    $\displaystyle\ell_{align}(\mathbf{f}_S, \mathbf{\hat{f}}_T^j) \defeq -\frac{1}{k} \sum_{i=1}^k \mathbbm{1}_{i=j} \left(\log \left(\operatorname{softmax}(\mathbf{\hat{s}}^i)_j\right) \right).$
    }
\end{align}

The above approach is a modification of a temporal cycle confusion objective proposed for robust object detection~\citep{wang2021robust}. 
However, we differ in three ways. First, we align cross-domain instance features between source and target domains, whereas the original approach aligns instance features across time given video data.
Second, we leverage target labels to align ROI features corresponding to \emph{ground truth} rather than predicted  bounding box coordinates.
Finally, our alignment objective uses cycle \emph{consistency} rather than cycle confusion. Intuitively, we encourage \emph{similar-looking} instances to be close together (by taking the negative Euclidean distance), whereas the original aligns dissimilar instances. 
Our alignment loss reduces to the classification of the soft nearest neighbors and therefore tends to be robust to label noise \cite{dwibedi2019temporal}.

\subsubsection{Bridging content gap with importance reweighting}

To close the task label distribution content gap, we apply inverse frequency reweighing to simulate a balanced label distribution in the source and target domains.
For each domain $\omega \in \{\source, \target\}$, we reweigh instances of class $C$ via multiplicative class weights $w_\omega(C) \propto 1/N_\omega(C)$, where $N_\omega(C)$ is the number of training examples in domain $\omega$. 

We approximate the class-conditional box ratios as follows
\begin{align}
   \frac{P_\target(B| C)}{P_\source(B| C)} 
    \approx \frac{P_\target(\wsf, \hsf | C)}{P_\source(\wsf, \hsf | C)} \frac{P_\target(\xsf, \ysf | C)}{P_\source(\xsf, \ysf | C)} =: v(B|C) \label{eq:wbc}
\end{align}
Intuitively, this ratio upweighs boxes of a class that are of a size and location relatively more represented in the target than in the source. 
Note that the first approximate equality $\approx$ is due to an assumption of independence between $(\wsf, \hsf)$ and $(\xsf, \ysf)$, which we assume to simplify computations.
We estimate each probability component via class-conditional Gaussian kernel density estimation (KDE) \citep{scott2015multivariate} fitted to the ground truth bounding box locations and sizes respectively. 
In Appendix \ref{sec:appendix_importance_reweighting}, we include details of this estimation, including appropriate smoothing and thresholding to handle regions with low target support.

\subsection{Analytical justification} 
\label{subsec:theoretical_interpretation}

We now analyze our loss function in Eq. \eqref{eq:dsda} to develop a theoretical intuition for its effectiveness. 
Let us rewrite the first term in the loss as follows:
\begin{align} \nonumber
    & \EX_{P_\source} \bigg[ w_\source(C) v(B|C) \ell_{det}(h(g(x)), B, C)\bigg] \\
    \nonumber
    =& \EX_{P_\target} \bigg[ \frac{P_\source(x, B, C)}{P_\target(x, B, C)} w_\source(C) v(B|C) \ell_{det}(h(g(x)), B, C) \bigg] \\
    \label{eq:expanded_first_term}
    \begin{split}
    =& \EX_{P_\target} \bigg[ \frac{P_\source(C)}{P_\target(C)} \times \frac{P_\source(B|C)}{P_\target(B|C)} \times \frac{P_\source(x|B,C)}{P_\target(x|B,C)} \\
    & \qquad\quad \times w_\source(C) v(B|C) \ell_{det}(h(g(x)), B, C)   \bigg].
    \end{split}
\end{align}
Above, the second line follows from importance reweighting, and the third line follows from Bayes rule.
Next, recall that $w_\source(C) = 1 / P_\source(C)$ and $v(B|C) \approx P_\target(B|C)/P_\source(B|C)$. Substituting these two, we obtain
\begin{align} \label{eq:simplified_first_term}
    \resizebox{0.91\hsize}{!}{
    $\displaystyle\text{Eq. }\eqref{eq:expanded_first_term} \approx  \EX_{P_\target} \bigg[ \frac{P_\source(x|B,C)}{P_\target(x|B,C)} \frac{1}{P_\target(C)} \ell_{det}(h(g(x)), B, C)  \bigg]$
    }
\end{align}

Finally, recall our feature alignment component, which is designed to minimize the distance between encoded features of the same class and box statistics. Successfully minimizing the third term in Eq. \eqref{eq:dsda} should obtain $P_\source(g(x) | B, C) = P_\target(g(x)| B, C)$. Using this, we obtain
\begin{align} \nonumber
    \text{Eq. }\eqref{eq:simplified_first_term} & \approx \EX_{P_\target} \bigg[ \frac{P_\source(g(x)|B,C)}{P_\target(g(x)|B,C)} \frac{1}{P_\target(C)} \ell_{det}(h(g(x)), B, C)  \bigg]  \\
    \label{eq:simplified_first_term_v2}
    & = \EX_{P_\target} \bigg[ \frac{1}{P_\target(C)} \ell_{det}(h(g(x)), B, C)  \bigg] 
\end{align}
where the first line follows from the assumption that feature-level distances should reflect image appearance distances, and the second line follows from minimizing $\ell_{align}$.
Overall, Eq. \eqref{eq:simplified_first_term_v2} and the second term in Eq. \eqref{eq:dsda} minimize a class-weighted version of the expected risk $r_\target$.
In our case, the target metric is mean AP, which values performance on all classes equally. Since in practice, our target data distributions often feature imbalanced classes, this modified risk simulates a balanced class label distribution and better maximizes mAP.

We remark that the steps here follow from several assumptions including independence of box position and size, equivalence between the ratios of feature-level probabilities and appearance probabilities, 
and that the target support is a subset of that of the source. Further, it relies on successfully minimizing this feature-level gap. Nonetheless as we show in the next section, our method demonstrates powerful empirical performance in the target domain.

%% file: sections/results.tex
\section{Results}
\label{ref:results}

\par\noindent We now describe our experimental setup for object detection adaptation: datasets and metrics (Section~\ref{subsec:datasets_det}), implementation details (Section \ref{subsec:implementation_det}), and baselines (Section~\ref{subsec:baselines_det}). We then present our results (Section~\ref{subsec:results_det}) and ablate (Section~\ref{subsec:ablations}) and analyze our approach (Section~\ref{subsec:analysis_det}).

\subsection{Datasets and metrics}
\label{subsec:datasets_det}

\begin{table*}[!t]    
  \centering
  \RawFloats \caption{\label{tab:results}Results for supervised sim2real object detection adaptation on target. We compare \method to source and target only training, a state-of-the-art unsupervised DA method (ILLUME~\cite{khindkar2022miss}), naive sim+real combinations (mixing~\cite{kishore2021synthetic} and sequential finetuning~\cite{tremblay2018training}), supervised extensions of popular UDA methods (DANN~\cite{ganin2014unsupervised} and MMD~\cite{long2015learning}),and a recently proposed few-shot detection strategy~\cite{wang2020frustratingly}.}
  \centering
  \begin{subfloatrow} 
  \hspace{0.5cm}
  \ffigbox[\FBwidth][][b]    
  {
  \footnotesize
      \begin{tabular}{lc}
        \toprule
         \multicolumn{1}{c}{\centering \bf Method} & \multicolumn{1}{c}{\centering \bf mAP@50 ($\uparrow$)} \\
        \toprule
        Source & 41.8 \\
        UDA  & 53.1 \\
        Target & 62.1 \\
        Mixing & 64.8 \\
        Seq. FT & 66.4 \\
        S-MMD &  65.8 \\
        S-DANN &  65.3 \\
        FDA &  65.2 \\
        \rowcolor{Gray}
        \method (ours) & \textbf{68.1} \\
        \bottomrule
        \end{tabular}
}
{
    \caption{\label{tab:ablate_com_att}\small Sim10K$\to$Cityscapes (1 class)}
    
  }
\ffigbox[\FBwidth][][!htbp]
{
  \footnotesize
      \begin{tabular}{lc}
        \toprule
         \multicolumn{1}{c}{\centering \bf Method} & \multicolumn{1}{c}{\centering \bf mAP@50 ($\uparrow$)~~~~} \\
        \toprule
        Source & 19.2 \\
        Target & 34.2 \\      
        Mixing & 39.0 \\
        Seq. FT & 39.8 \\
        S-MMD & 40.0 \\
        S-DANN & 40.8 \\
        \rowcolor{Gray}
        \method (ours) & \textbf{48.5} \\      
        \bottomrule
        \end{tabular}
}
  {
  \caption{\label{tab:ablate_mr}\small Synscapes$\to$Cityscapes (8 classes)}
  
  } \hspace{0.5cm}
  \ffigbox[\FBwidth][][!htbp]
{
  \footnotesize
      \begin{tabular}{lc}
        \toprule
         \multicolumn{1}{c}{\centering \bf Method} & \multicolumn{1}{c}{\centering \bf mAP@50 ($\uparrow$)~~~~~~~~~~~~~} \\
         \toprule
        Source & 22.5 \\
      Target & 45.2 \\      
      Mixing & 49.3 \\
      Seq. FT & 45.4 \\
      S-MMD &  50.6\\
      S-DANN & 49.8 \\
      \rowcolor{Gray}
      \method (ours) & \textbf{53.7} \\        
        \bottomrule
        \end{tabular}
}
  {
  \caption{\label{tab:ablate_mr}\small \drivesimtocityscapes (3 classes)}
  
  }  
  \end{subfloatrow}  
\end{table*}

\noindent We perform domain adaptation from three different source data sets of synthetic images, Sim10K \citep{johnson2017driving}, Synscapes \citep{wrenninge2018synscapes}, and \drivesim, an internal data set simulated using an early version of NVIDIA DriveSim ~\cite{nvdrivesim}, following the procedure described in ~\citet{acuna2021towards}.
Sim10K contains 10,000 images of 1914$\times$1052 resolution with pixel-level annotations extracted from the game GTA-5.
Synscapes is a photorealistic dataset of 25,000 synthetic driving scenes of 1440$\times$720 resolution.
Finally, \drivesim is a private synthetic data set of 48,000 photorealistic driving scenes. 
Synscapes and \drivesim exhibit a long-tailed category distribution (see Fig.~\ref{fig:domain_gap}).
For each source, we train an object detector to adapt to our target, Cityscapes \citep{cordts2016cityscapes} which is a data set of 2500 real driving images. For all evaluations, we fix the target data set size to 25\% to model the realistic scenario of available but an order of magnitude less real data than synthetic  data (see appendix for details).
For \textbf{Sim10K$\to$Cityscapes}, we focus on object detection for a single class (\ie car) to better compare against prior Sim2Real domain adaptation methods \citep{khindkar2022miss}. 
For \textbf{Synscapes$\to$Cityscapes} and \textbf{\drivesimtocityscapes}, we evaluate object detection for eight and three classes, respectively.
To evaluate all models, we match prior work~\cite{chen2018domain, khindkar2022miss,wang2021robust} and report per-category Average Precision (AP) and its mean across classes at an IoU threshold of 50\% (mAP@50), over the target test set.

\subsection{Implementation details}
\label{subsec:implementation_det}

We use a Faster-RCNN~\cite{ren2015faster} architecture with a ResNet-50~\cite{he2016deep} backbone. 
We run 10k iterations of SGD with a learning rate of 0.01, momentum of 0.9, weight decay of 10$^{-4}$, and learning rate warmup matching ~\cite{wang2021robust}.
We set $\lambda=0.1$ in Eq. \eqref{eq:dsda}. We use 8 NVIDIA V100 GPUs with a per-GPU batch size of 4, and maintain a 1:1 within-batch source to target ratio across experiments.

\subsection{Baselines}
\label{subsec:baselines_det}

We compare against: 
(1) \textbf{Source only}: Supervised learning using only the labeled source dataset.
(2) \textbf{Target only}: Supervised learning using only the labeled target dataset.
(3) \textbf{Mixing}~\cite{kishore2021synthetic}: Supervised learning on the combined source and target data sets, while maintaining a 1:1 ratio within batches (we ablate this mixing ratio in appendix). 
(4) \textbf{Sequential Finetuning}~\cite{tremblay2018training}: Supervised learning on the source dataset followed by finetuning all layers of the model with the target dataset. 
(5) \textbf{Unsupervised DA (UDA) with ILLUME}~\cite{khindkar2022miss}: For completeness, we copy results on Sim10K$\to$Cityscapes of a state-of-the-art UDA method that uses labeled source and unlabeled target data.

We also propose and benchmark supervised extensions of two popular UDA strategies:
(6) \textbf{S-MMD}: A class and box-conditional \emph{supervised} version of Maximum Mean Discrepancy~\cite{long2015learning}. S-MMD minimizes the MMD loss between cross-domain box features corresponding to the same class, using a linear kernel. 
(7) \textbf{S-DANN}: A class and box-conditional \emph{supervised} version of DANN~\cite{ganin2014unsupervised}. S-DANN minimizes the domain adversarial loss between cross-domain box features corresponding to the same class, similar to ~\citet{chen2018domain}. (8) \textbf{Few-shot DA (FDA) with TFA:}~\cite{wang2020frustratingly}. This is a two-stage finetuning algorithm proposed for few-shot object detection that updates all parameters on source (base) data followed by finetuning only the final layer (box regressor and classifer) on a balanced dataset of source and target data. However, we observe low performance with finetuning only the last layer (despite using a lower learning rate as recommended and both with and without weight re-initialization). Instead, we report results \emph{without} freezing weights in the second phase.

\subsection{Main Results}%
\label{subsec:results_det}

Table \ref{tab:results} summarizes our results. We find:

\begin{table*}[!ht] 
  \centering 
  \RawFloats  \caption{\label{tab:ablations} Ablating our proposed method on all three shifts. Our method is in gray with the improvement versus mixing in small font.}
  \centering
  \footnotesize
  \begin{tabular}{llcccccc} 
      \toprule
      \multirow{2}{*}{\#} &\multicolumn{1}{c}{\centering  $P(g(x)|B, C)$}&  \multicolumn{1}{c}{\centering $P(C)$ } & \multicolumn{1}{c}{\centering  $P(B|C)$}  & \multicolumn{3}{c}{\centering mAP@50 ($\uparrow$)} \\
      \cmidrule(l{4pt}r{4pt}){5-7}
      &\multicolumn{1}{c}{\centering alignment} & \multicolumn{1}{c}{\centering rewt.}  & \multicolumn{1}{c}{\centering rewt.} & {\centering Sim10k }& {\centering Synscapes } & {\centering \drivesim} \\ 
      \toprule                 
      1&\multicolumn{3}{c}{\multirow{1}{*}{  \centering (Mixing baseline)}} & 64.8 &39.0 & 49.3  \\
      \midrule         
      2&S-MMD &  &  & 65.8  & 40.0& 50.6   \\
      3&S-DANN &  &  & 65.3  & 40.8& 49.8   \\           
      4&Cycle Consistency &  &  & 67.2  & 41.8& 50.8    \\
      5& None (Mixing baseline) & \ding{51} &  &  64.8  & 46.1 & 51.8 \\                          
       6&Cycle Consistency & \ding{51} &  &  67.2 & 46.6 & 52.5      \\              
      \midrule
      \rowcolor{Gray}
      7&Cycle Consistency& \ding{51} & \ding{51} & \textbf{68.1}\scriptsize{+3.3}  & \textbf{48.5}\scriptsize{+9.5}  & \textbf{53.7}\scriptsize{+4.4}      \\
      \bottomrule
      \end{tabular}
\end{table*}

\noindent\textbf{$\triangleright$ Simulated data and labeled real data are both needed.}
We first confirm that supervised learning using only the target data outperforms
both the settings of using only source data and unsupervised domain adaptation with unlabeled target data.
Moreover across all three shifts, even baselines that na\"ively combine simulated and real data (\ie mixing and sequential finetuning) outperform training using only the target data.
This shows that additional simulated data is helpful. Moreover, sequential finetuning outperforms mixing on two of three shifts. Finally, we find that mixing with additional conditional feature alignment (S-MMD, S-DANN), consistently outperforms naive mixing. Additional results are in Appendix~\ref{subsec:doweneed}.

\noindent\textbf{$\triangleright$ \method outperforms all competing methods.} First, note that across each shift, \method outperforms mixing (\textbf{+3.3}, \textbf{+9.5}, \textbf{+4.4} mAP@50) and sequential finetuning (\textbf{+1.7}, \textbf{+8.7}, \textbf{+8.3} mAP@50). This suggests that the Sim2Real domain gap is a barrier to effective mixing, and systematically mitigating it using target labels is beneficial.
Most importantly, we outperform each benchmarked supervised extension of UDA on all shifts. 
This result validates the research-practice gap by showing that UDA cannot be easily extended to the practical setting of labeled target data, thereby necessitating \method in supervised domain adaptation.

\subsection{Ablation study}
\label{subsec:ablations}

In Table~\ref{tab:ablations}, we ablate the various components of \method. %

\noindent\textbf{$\triangleright$ Class-and-box conditional feature alignment is necessary (Rows 2-4 vs. 1).} Regardless of the specific feature alignment strategy (\ie S-MMD, S-DANN, and our proposed cross-domain Cycle Consistency), additional feature alignment improves performance. 

We also remark that during model design, we tested variations of Cycle Consistency-based alignment on Sim10K$\to$Cityscapes by i) conditioning on \emph{predicted} rather than ground truth class and box coordinates (66.1 mAP@50, \textbf{-1.1} compared to our method), and ii) conditioning on predicted box coordinates and ignoring class predictions (64.9 mAP@50, roughly on par with mixing). These two settings yielded 66.1 mAP@50 (\textbf{-1.1} versus Row 4) and 64.9 mAP@50 (\textbf{-2.3} versus Row 4), respectively. 
Finally, we also tested a dissimilarity variant of our approach (\ie similar to \citet{wang2021robust}) instead of consistency matching for feature alignment. This approach performs on par with Row 4 (67.3 mAP@50 on Sim10K$\to$Cityscapes), and we consequently opted to keep cycle consistency throughout.

\begin{table}[!t]    
  \centering
  \RawFloats 
  \caption{\label{tab:ablations2} \small Ablating our proposed conditional reweighting strategies on Synscapes $\to$ Cityscapes.} 
  \begin{subfloatrow}  \footnotesize
  \ffigbox[\FBwidth][][b]    
  {
      \setlength{\tabcolsep}{2pt}
      \begin{tabular}{lc c}
        \toprule
        \textbf{Method} & \textbf{w/o CB} & \textbf{w/ CB} \\
        \midrule
        Source & 19.2 & 20.0 \\                  
        Target & 34.2 & 40.0\\                  
        Mixing & 39.0 & 46.1 \\
        Seq. FT & 39.8 & 44.9 \\              
        \bottomrule
        \end{tabular}
}
{
    \caption{\small Ablating $P(C)$ rewt.}
    \label{tab:ablate_com_att}
  }
\hspace{-0.5cm}
\ffigbox[\FBwidth][][!htbp]
{
      \setlength{\tabcolsep}{2pt}
      \begin{tabular}{l c}
        \toprule
        \textbf{Method} & \textbf{mAP@50} \\
        \midrule
        \rowcolor{Gray}
        $P(\wsf, \hsf, \xsf, \ysf | C)$  & 48.5 \\          
        Only $P(\xsf, \ysf | C)$ & 46.7\\        
        Only $P(\wsf, \hsf | C)$ & 48.3 \\     
        None & 46.6 \\          
        \bottomrule
        \end{tabular}
}
  {
  \caption{\small Ablating $P(B|C)$ rewt.}
  \label{tab:ablate_mr}
  }
  \end{subfloatrow}
\end{table}

\begin{figure}[t]
 \centering  \includegraphics[width=0.9\linewidth]{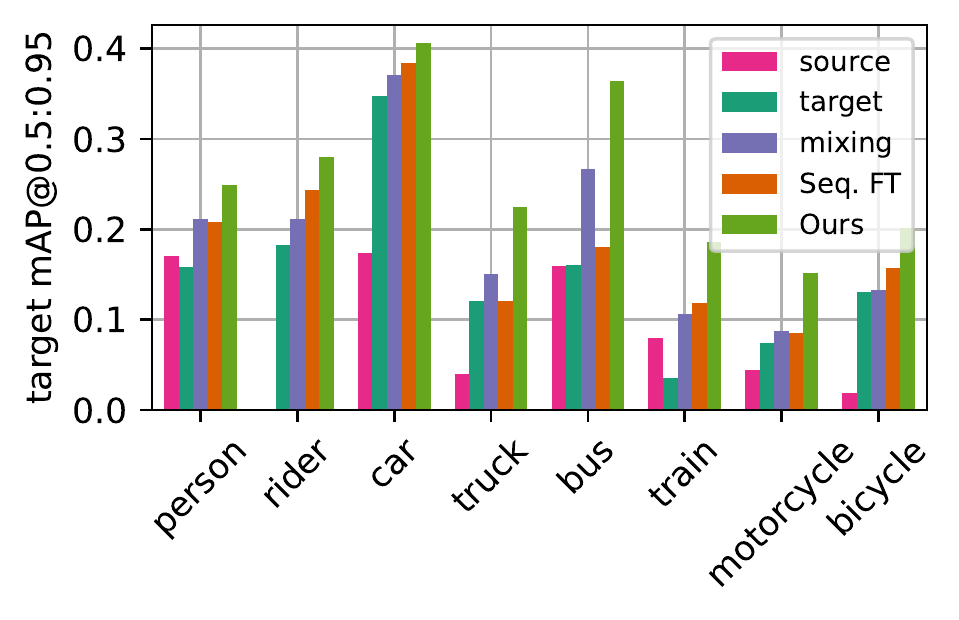}
     \caption[]%
     {Per-class performance comparison of \method to baselines on Synscapes$\to$Cityscapes.}   
   \label{fig:perclass}
\end{figure}

\subsection{\method: Fine-grained performance analysis}
\label{subsec:analysis_det}

\begin{figure}[t]
 \centering
 \begin{subfigure}[t]{1\textwidth}
 \centering
 \includegraphics[width=0.7\textwidth]{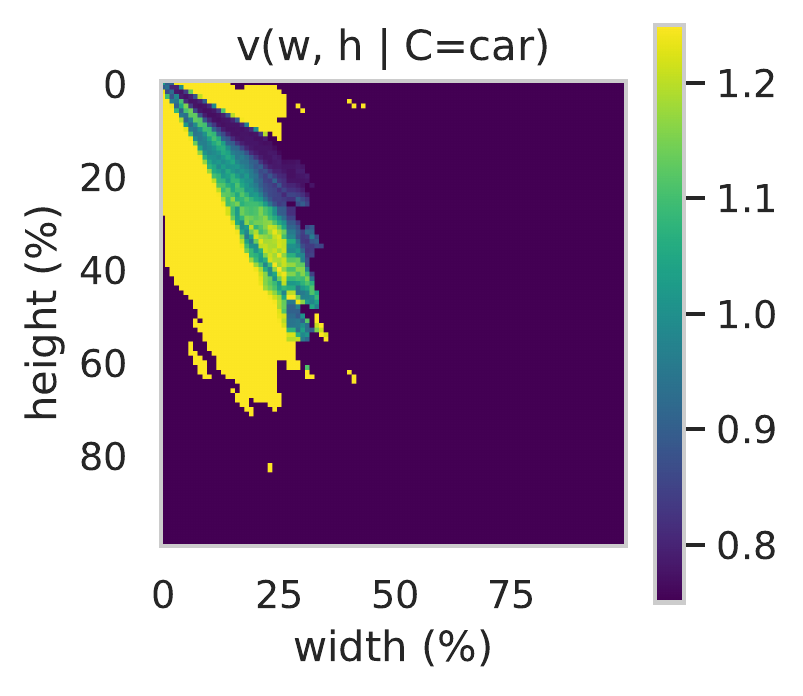}
 \end{subfigure}
 \begin{subfigure}[t]{1\textwidth}
 \includegraphics[width=\textwidth]{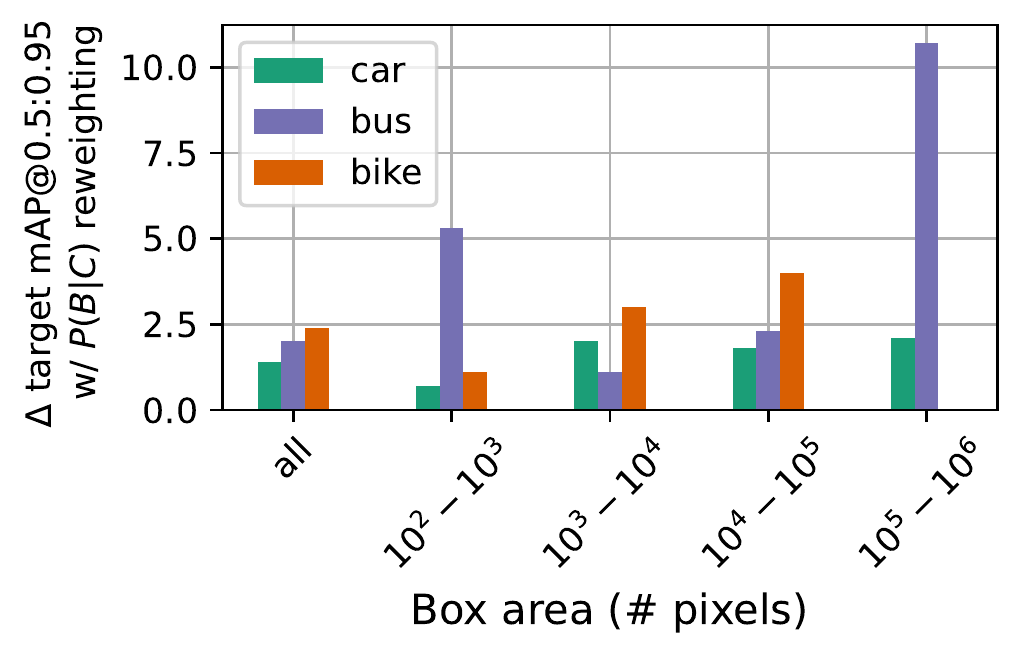}
 \end{subfigure}
 \caption{\label{fig:pbc}Visualizing $P(\wsf, \hsf|C)$ reweighting on Synscapes$\to$Cityscapes. \textbf{(top)} Visualizing $v(\wsf, \hsf|C=\text{car})$. \textbf{(bottom)} Visualizing change in mAP after $P(\wsf, \hsf|C)$ reweighting for three categories (car, bus, bike).}
 
\end{figure}

\begin{figure}[t]
 \centering
 \includegraphics[width=0.9\textwidth]{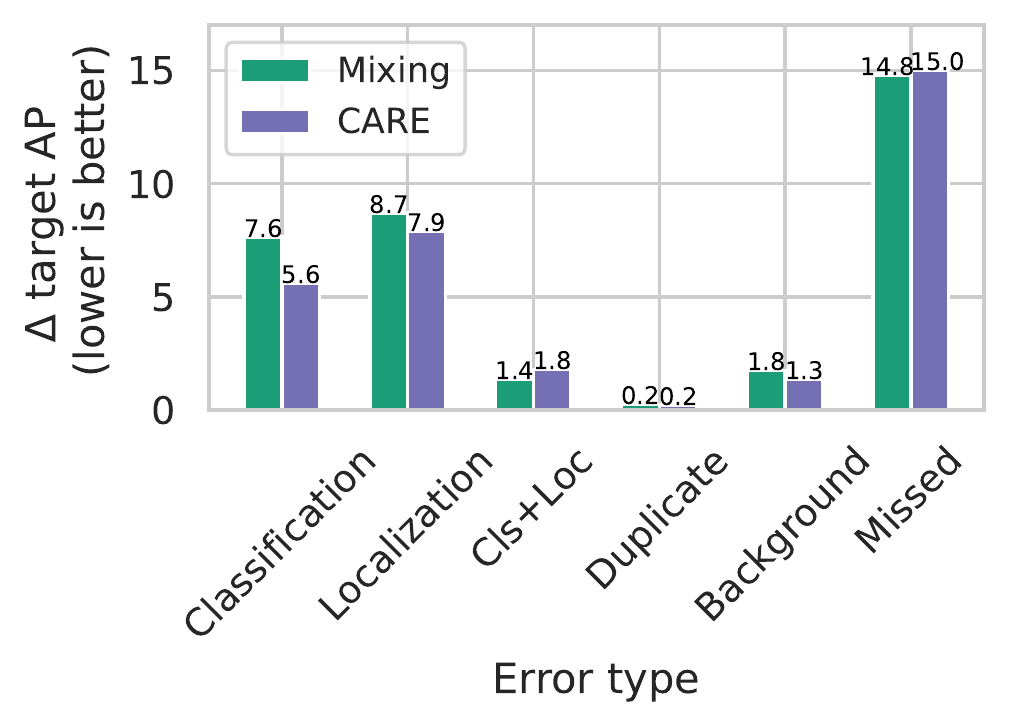}
 \caption{
   Visualizing change in dAP (lower is better)~\cite{bolya2020tide} for errors of different types using \method, over a mixing baseline.
 }
 \label{fig:tide}
 \vspace{-10pt}
\end{figure}

\noindent\textbf{$\triangleright$ $P(C)$ reweighting is highly effective (Row 5 vs. 1).} Particularly on multi-class source settings (\emph{e.g.} Synscapes and \drivesim), $P(C)$ reweighting considerably boosts performance. Further, Table~\ref{tab:ablations2} (a) shows that class balancing naturally improves the baselines as well, due to mAP evaluating classes equally. %

\noindent\textbf{$\triangleright$ $P(B|C)$ reweighting is helpful (Row 7 vs. 1).} Finally, we show that additional class-conditional box reweighting consistently improves performance across all shifts. Table~\ref{tab:ablations2} (b) presents results for different formulations of $P(B|C)$. It validates our reweighing scheme which decomposes box size with $P(\wsf, \hsf | C)$ and location with $P(\xsf, \ysf | C)$. Capturing both components is better than using only one or neither.

Using Synscapes$\to$Cityscapes, we analyze content-specific metrics to demonstrate \method consistently outperforms baselines on all settings and not just in aggregate.

\noindent\textbf{$\triangleright$ \method improves over baselines on all classes.} 
Fig.~\ref{fig:perclass} studies per-class performance improvements with our proposed method against baselines. Our method outperforms each baseline for every class. %

\noindent\textbf{$\triangleright$ \method improves per-class performance across box sizes.} 
Fig.~\ref{fig:pbc} (\emph{top}) visualizes bounding box frequency ratio weights $v(\wsf, \hsf| C)$ for the ``car'' class estimated via the first term of Eq.~\eqref{eq:wbc}. Matching our intuition (see Fig.~\ref{fig:domain_gap}, \emph{right}), these ratios upweigh target cars of sizes that are relatively less frequent in the source domain. Fig.~\ref{fig:pbc} (\emph{bottom}) illustrates the change in mAP as a result of our reweighing for three categories over boxes of different sizes. Here, reweighing consistently improves mAP and can yield up to $+10$ mAP improvement for large objects such as buses. We remark that these trends also hold for the remaining categories.

\noindent\textbf{$\triangleright$ Fine-grained error analysis.} We use the TIDE~\cite{bolya2020tide} toolbox to evaluate specific error types of our mixing baseline and \method models (lower is better). Fig.~\ref{fig:tide} shows that \method reduces classification, localization, and duplicate errors, while slightly worsening joint classification+localization errors.

\noindent\textbf{$\triangleright$ Visualizing matching with cycle consistency.} 
Fig.~\ref{fig:viz} provides a qualitative visualization of the matching behavior of our proposed cycle consistency approach, for two pairs of source and target images. For each example, we estimate the Euclidean distance in feature space between all cross-domain instance pairs in the aligned feature space of our \method model and visualize the closest pair of car instances for each example. As expected, we find that our method embeds similar looking cars closer in feature space.

%% file: sections/conclusion.tex
\vspace{-10pt}
\section{Discussion}
\label{ref:conclusion}

We study supervised Sim2Real adaptation applied to object detection, and propose a strategy that exploits target labels to explicitly estimate and bridge the sim2real appearance and content gaps. 
Our method possesses a clear theoretical intuition and our empirical analyses validate our improvements in every setting that we tested, for example by boosting mAP@50 by as much as $\sim$25\%. 
Most importantly, this paper tackles a large research-practice gap by bridging the literature on unsupervised and few-shot domain adaptation with an industry-standard practice of combining labeled data from both simulated and real domains. 
With this, we envision a renewed future methodological interest in SDA.

\textbf{Limitations.}
Our method requires sufficient labeled data in source and target domains to reliably estimate dataset-level statistics. Further, our formulation assumes conditional independence of box sizes and locations as well as an equivalence between pixel-level and feature-level distributions. We also rely on successful cross-domain alignment. These assumptions may be violated to varying degrees in practice. We  focus on object detection and the applicability of our method to other tasks, while plausible, is not established. Finally, we do not consider an unlabeled portion of the target domain and leave that exploration to future work.

%% file: sections/appendix.tex
\section{Appendix}
\label{sec:appendix}

\subsection{When is adding simulated data most helpful?}
\label{subsec:doweneed}

\begin{figure}[t]
  \centering
  \includegraphics[width=0.5\textwidth]{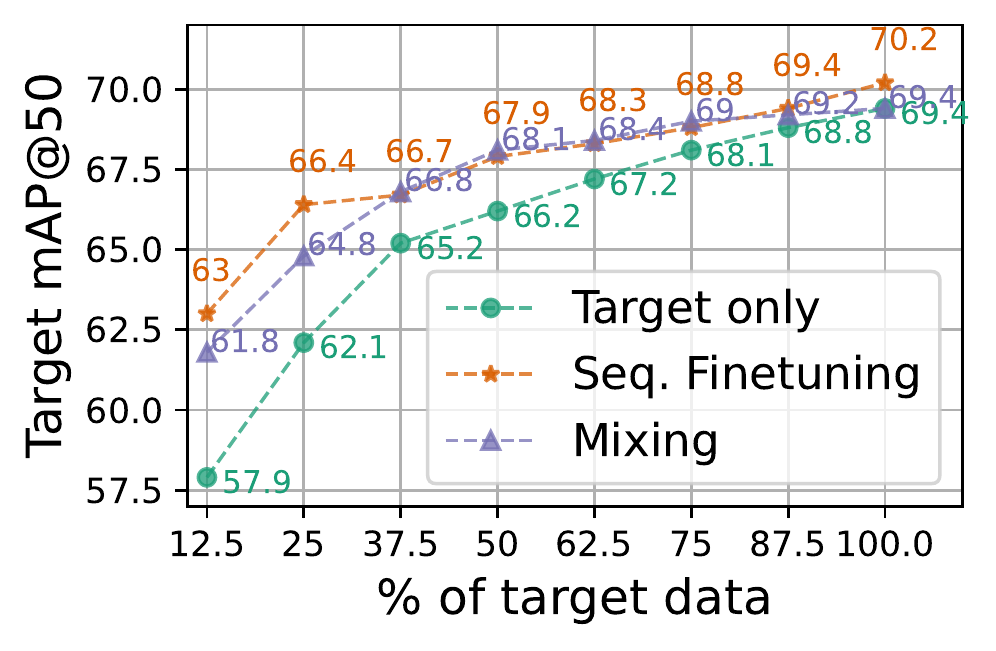}
  \caption{
  Plotting the scaling behavior of mixing, seq. finetuning, and target only baselines using 100\% of source data and a varying amount of target data, on Sim10K$\to$Cityscapes.
  }
  \label{fig:scaling}
  \vspace{-5pt}
\end{figure}

Intuitively, collecting real instances corresponding to the long-tail of driving scenarios is extremely challenging, and simulated data can offer inexpensive, unlimited labeled data to augment our dataset, albeit with an additional domain gap. However the utility of such additional data may depend on the nature and amount of labeled target available. A natural question is then: in what situations can additional simulated data help boost target performance? 

To study this, we benchmark simple baselines on the Sim10K$\to$Cityscapes shift for car detection. We plot the performance of training using only target data, mixing, and sequential finetuning strategies as we vary the amount of target data. For mixing and sequential finetuning we additionally use 100\% of source data. As Figure~\ref{fig:scaling} shows, both baselines improve upon target only training, with sequential finetuning initially outperforming mixing. However, with the relatively small target task that we study, gains over target-only training are clearly more pronounced in the low target data regime (\textbf{+4.3} mAP@50 at 25\% with Seq. FT), and performance saturates as we acquire more labeled target data. For Sim10K$\to$Cityscapes, atleast with naive combinations of simulated and real data, we find that adding simulated data has maximum relative utility in the limited target label regime and diminishing returns thereafter. 
However, even performance with target-only training saturates towards the end, and it is unclear if the diminishing returns are a consequence of that. Further, it is unclear whether more principled combinations of sim and real data (via \method) will exhibit similar trends. Nevertheless, to rigorously study the supervised DA setting we reduce the number of target data points (in our experiments, we randomly subsample 25\%$\sim$625 examples of the target train dataset and use it for adaptation, while leaving the test set unchanged).

\begin{figure}[t]
  \centering  
      \includegraphics[width=0.5\linewidth]{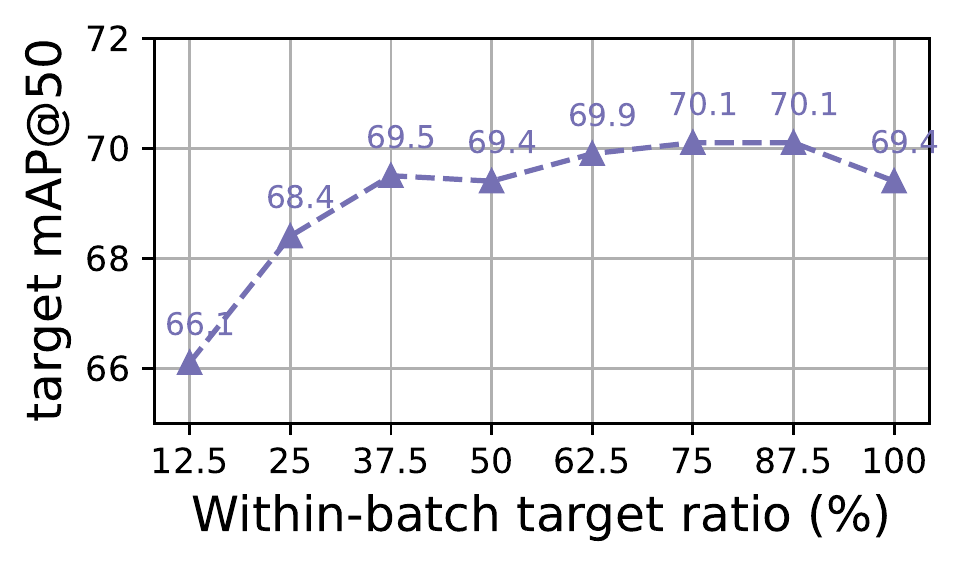}
      \caption[]%
      {{\small Sim10K$\to$Cityscapes (100\% target data): Varying within-batch real:sim ratio for mixing.}}  
    \label{fig:mixing_ratio}
\end{figure}

\noindent\textbf{$\triangleright$ Varying mixing ratio.} The mixing ratio controls the proportion of source and target data that are contained within a minibatch of fixed size. In Fig.~\ref{fig:mixing_ratio}, we vary the mixing ratio of real:sim data and measure the subsequent performance on the Cityscapes test set. We use all source and target data for these experiments. We observed that the performance is fairly stable beyond a ratio of 50\% and, for simplicity, we adopt this mixing ratio for all experiments unless otherwise stated.

\subsection{Additional details on Importance reweighting} \label{sec:appendix_importance_reweighting}

When applying bounding-box importance reweighting we introduce a smoothing mechanism to ensure bounded loss values and thresholding to handle areas with low target support. Specifically, we compute:
$$
v\left(B \mid C\right)= \begin{cases}\alpha  \sigma\left(\frac{P_T\left(B \mid C\right)}{P_S\left(B \mid C\right)} \right)+\beta & \text { if } P_T\left(B \mid C\right)>\tau \\ 1.0 & \text { otherwise }\end{cases}
$$
where $\alpha, \beta$ are scaling parameters (that we set to 20, -9, effectively bounding loss weights between 1 and 11). $\sigma$ denotes the sigmoid operator, and $\tau$ is a probability threshold that we set to 0.1. For boxes with very small target support, we simply set weights to a floor value of 1.0.

\subsection{Additional \method analysis: Visualizing KDE estimates} 
\begin{figure}[t]
  \centering
  \includegraphics[width=0.6\textwidth]{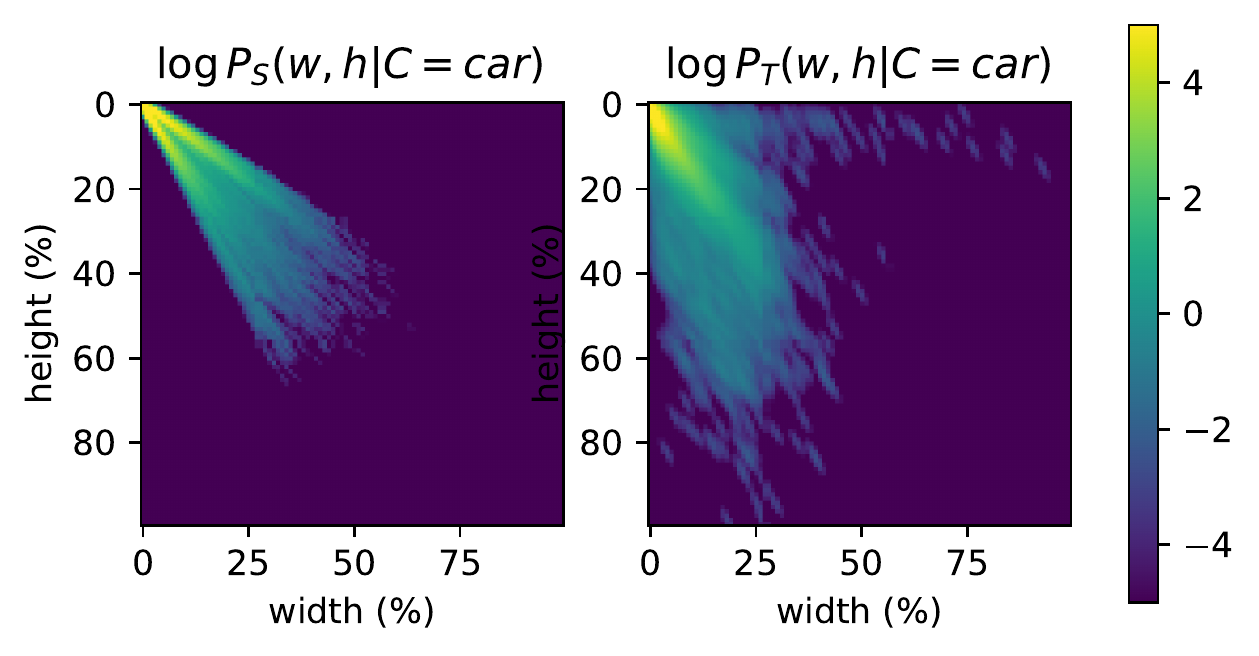}
  \caption{
  Visualizing log PDF values of KDE densities fitted to bounding box size on Synscapes$\to$Cityscapes for the ``car'' class.
  }
  \label{fig:kdeviz}
\end{figure}

In Figure \ref{fig:kdeviz}, we visualize log PDF values from KDE estimates fit to bounding box width and heights on both the source (Synscapes) and target (Cityscapes) domains. As seen, the KDE is able to capture the difference in box size distributions across domains: car class sizes vary significantly more in the target domain, consistent with our observation in Fig.~\ref{fig:domain_gap}. It is intuitive that with appropriate importance reweighting for source boxes as proposed in Sec.~\ref{sec:method}, we can improve overall performance across categories.

%% file: main.bbl
\begin{thebibliography}{48}
\providecommand{\natexlab}[1]{#1}
\providecommand{\url}[1]{\texttt{#1}}
\expandafter\ifx\csname urlstyle\endcsname\relax
  \providecommand{\doi}[1]{doi: #1}\else
  \providecommand{\doi}{doi: \begingroup \urlstyle{rm}\Url}\fi

\bibitem[Acuna et~al.(2021{\natexlab{a}})Acuna, Philion, and
  Fidler]{acuna2021towards}
Acuna, D., Philion, J., and Fidler, S.
\newblock Towards optimal strategies for training self-driving perception
  models in simulation.
\newblock In Beygelzimer, A., Dauphin, Y., Liang, P., and Vaughan, J.~W.
  (eds.), \emph{Advances in Neural Information Processing Systems},
  2021{\natexlab{a}}.
\newblock URL \url{https://openreview.net/forum?id=ZfIO21FYv4}.

\bibitem[Acuna et~al.(2021{\natexlab{b}})Acuna, Zhang, Law, and
  Fidler]{pmlr-v139-acuna21a}
Acuna, D., Zhang, G., Law, M.~T., and Fidler, S.
\newblock f-domain adversarial learning: Theory and algorithms.
\newblock In Meila, M. and Zhang, T. (eds.), \emph{Proceedings of the 38th
  International Conference on Machine Learning}, volume 139 of
  \emph{Proceedings of Machine Learning Research}, pp.\  66--75. PMLR, 18--24
  Jul 2021{\natexlab{b}}.

\bibitem[Bolya et~al.(2020)Bolya, Foley, Hays, and Hoffman]{bolya2020tide}
Bolya, D., Foley, S., Hays, J., and Hoffman, J.
\newblock Tide: A general toolbox for identifying object detection errors.
\newblock In \emph{Computer Vision--ECCV 2020: 16th European Conference,
  Glasgow, UK, August 23--28, 2020, Proceedings, Part III 16}, pp.\  558--573.
  Springer, 2020.

\bibitem[Chattopadhyay et~al.(2022)Chattopadhyay, Sarangmath, Vijaykumar, and
  Hoffman]{chattopadhyay2022pasta}
Chattopadhyay, P., Sarangmath, K., Vijaykumar, V., and Hoffman, J.
\newblock Pasta: Proportional amplitude spectrum training augmentation for
  syn-to-real domain generalization.
\newblock \emph{arXiv preprint arXiv:2212.00979}, 2022.

\bibitem[Chen et~al.(2018)Chen, Li, Sakaridis, Dai, and
  Van~Gool]{chen2018domain}
Chen, Y., Li, W., Sakaridis, C., Dai, D., and Van~Gool, L.
\newblock Domain adaptive faster r-cnn for object detection in the wild.
\newblock In \emph{Proceedings of the IEEE conference on computer vision and
  pattern recognition}, pp.\  3339--3348, 2018.

\bibitem[Cordts et~al.(2016)Cordts, Omran, Ramos, Rehfeld, Enzweiler, Benenson,
  Franke, Roth, and Schiele]{cordts2016cityscapes}
Cordts, M., Omran, M., Ramos, S., Rehfeld, T., Enzweiler, M., Benenson, R.,
  Franke, U., Roth, S., and Schiele, B.
\newblock The cityscapes dataset for semantic urban scene understanding.
\newblock In \emph{Proceedings of the IEEE conference on computer vision and
  pattern recognition}, pp.\  3213--3223, 2016.

\bibitem[Csurka et~al.(2022)Csurka, Volpi, and Chidlovskii]{csurka2022}
Csurka, G., Volpi, R., and Chidlovskii, B.
\newblock Unsupervised domain adaptation for semantic image segmentation: a
  comprehensive survey.
\newblock \emph{Foundations and Trends in Computer Graphics and Vision}, 2022.

\bibitem[Donahue et~al.(2013)Donahue, Hoffman, Rodner, Saenko, and
  Darrell]{donahue2013semi}
Donahue, J., Hoffman, J., Rodner, E., Saenko, K., and Darrell, T.
\newblock Semi-supervised domain adaptation with instance constraints.
\newblock In \emph{Proceedings of the IEEE conference on computer vision and
  pattern recognition}, pp.\  668--675, 2013.

\bibitem[Dwibedi et~al.(2019)Dwibedi, Aytar, Tompson, Sermanet, and
  Zisserman]{dwibedi2019temporal}
Dwibedi, D., Aytar, Y., Tompson, J., Sermanet, P., and Zisserman, A.
\newblock Temporal cycle-consistency learning.
\newblock In \emph{Proceedings of the IEEE/CVF conference on computer vision
  and pattern recognition}, pp.\  1801--1810, 2019.

\bibitem[Ganin \& Lempitsky(2015)Ganin and Lempitsky]{ganin2014unsupervised}
Ganin, Y. and Lempitsky, V.
\newblock Unsupervised domain adaptation by backpropagation.
\newblock In \emph{International Conference on Machine Learning}, pp.\
  1180--1189, 2015.

\bibitem[Gao et~al.(2022)Gao, Yang, Huang, Xie, Li, and Zheng]{gao2022acrofod}
Gao, Y., Yang, L., Huang, Y., Xie, S., Li, S., and Zheng, W.-S.
\newblock Acrofod: An adaptive method for cross-domain few-shot object
  detection.
\newblock In \emph{Computer Vision--ECCV 2022: 17th European Conference, Tel
  Aviv, Israel, October 23--27, 2022, Proceedings, Part XXXIII}, pp.\
  673--690. Springer, 2022.

\bibitem[He et~al.(2016)He, Zhang, Ren, and Sun]{he2016deep}
He, K., Zhang, X., Ren, S., and Sun, J.
\newblock Deep residual learning for image recognition.
\newblock In \emph{Proceedings of the IEEE conference on computer vision and
  pattern recognition}, pp.\  770--778, 2016.

\bibitem[Hoffman et~al.(2013)Hoffman, Rodner, Donahue, Darrell, and
  Saenko]{hoffman2013efficient}
Hoffman, J., Rodner, E., Donahue, J., Darrell, T., and Saenko, K.
\newblock Efficient learning of domain-invariant image representations.
\newblock \emph{arXiv preprint arXiv:1301.3224}, 2013.

\bibitem[Hoffman et~al.(2018)Hoffman, Tzeng, Park, Zhu, Isola, Saenko, Efros,
  and Darrell]{hoffman2018cycada}
Hoffman, J., Tzeng, E., Park, T., Zhu, J.-Y., Isola, P., Saenko, K., Efros, A.,
  and Darrell, T.
\newblock Cycada: Cycle-consistent adversarial domain adaptation.
\newblock In \emph{International conference on machine learning}, pp.\
  1989--1998. Pmlr, 2018.

\bibitem[Johnson-Roberson et~al.(2017)Johnson-Roberson, Barto, Mehta, Sridhar,
  Rosaen, and Vasudevan]{johnson2017driving}
Johnson-Roberson, M., Barto, C., Mehta, R., Sridhar, S.~N., Rosaen, K., and
  Vasudevan, R.
\newblock Driving in the matrix: Can virtual worlds replace human-generated
  annotations for real world tasks?
\newblock In \emph{2017 IEEE International Conference on Robotics and
  Automation (ICRA)}, pp.\  746--753. IEEE, 2017.

\bibitem[Kar et~al.(2019)Kar, Prakash, Liu, Cameracci, Yuan, Rusiniak, Acuna,
  Torralba, and Fidler]{kar2019meta}
Kar, A., Prakash, A., Liu, M.-Y., Cameracci, E., Yuan, J., Rusiniak, M., Acuna,
  D., Torralba, A., and Fidler, S.
\newblock Meta-sim: Learning to generate synthetic datasets.
\newblock In \emph{Proceedings of the IEEE/CVF International Conference on
  Computer Vision}, pp.\  4551--4560, 2019.

\bibitem[Karpathy(2021)]{teslaaiday}
Karpathy, A.
\newblock Tesla ai day 2021 -simulation, 2021.
\newblock URL \url{https://www.youtube.com/watch?v=j0z4FweCy4M&t=5692s}.

\bibitem[Khindkar et~al.(2022)Khindkar, Arora, Balasubramanian, Subramanian,
  Saluja, and Jawahar]{khindkar2022miss}
Khindkar, V., Arora, C., Balasubramanian, V.~N., Subramanian, A., Saluja, R.,
  and Jawahar, C.
\newblock To miss-attend is to misalign! residual self-attentive feature
  alignment for adapting object detectors.
\newblock In \emph{Proceedings of the IEEE/CVF Winter Conference on
  Applications of Computer Vision}, pp.\  3632--3642, 2022.

\bibitem[Kim et~al.(2019)Kim, Choi, Kim, and Kim]{kim2019self}
Kim, S., Choi, J., Kim, T., and Kim, C.
\newblock Self-training and adversarial background regularization for
  unsupervised domain adaptive one-stage object detection.
\newblock In \emph{Proceedings of the IEEE/CVF International Conference on
  Computer Vision}, pp.\  6092--6101, 2019.

\bibitem[Kishore et~al.(2021)Kishore, Choe, Kwon, Park, Hao, and
  Mittel]{kishore2021synthetic}
Kishore, A., Choe, T.~E., Kwon, J., Park, M., Hao, P., and Mittel, A.
\newblock Synthetic data generation using imitation training.
\newblock In \emph{Proceedings of the IEEE/CVF International Conference on
  Computer Vision}, pp.\  3078--3086, 2021.

\bibitem[Kulis et~al.(2011)Kulis, Saenko, and Darrell]{kulis2011you}
Kulis, B., Saenko, K., and Darrell, T.
\newblock What you saw is not what you get: Domain adaptation using asymmetric
  kernel transforms.
\newblock In \emph{CVPR 2011}, pp.\  1785--1792. IEEE, 2011.

\bibitem[Li et~al.(2022)Li, Dai, Ma, Liu, Chen, Wu, He, Kitani, and
  Vajda]{li2022cross}
Li, Y.-J., Dai, X., Ma, C.-Y., Liu, Y.-C., Chen, K., Wu, B., He, Z., Kitani,
  K., and Vajda, P.
\newblock Cross-domain adaptive teacher for object detection.
\newblock In \emph{Proceedings of the IEEE/CVF Conference on Computer Vision
  and Pattern Recognition}, pp.\  7581--7590, 2022.

\bibitem[Long et~al.(2015)Long, Cao, Wang, and Jordan]{long2015learning}
Long, M., Cao, Y., Wang, J., and Jordan, M.
\newblock Learning transferable features with deep adaptation networks.
\newblock In \emph{International Conference on Machine Learning}, pp.\
  97--105, 2015.

\bibitem[Long et~al.(2018)Long, Cao, Wang, and Jordan]{long2018conditional}
Long, M., Cao, Z., Wang, J., and Jordan, M.~I.
\newblock Conditional adversarial domain adaptation.
\newblock \emph{Advances in neural information processing systems}, 31, 2018.

\bibitem[Mahmood et~al.(2022)Mahmood, Lucas, Alvarez, Fidler, and
  Law]{mahmood2022optimizing}
Mahmood, R., Lucas, J., Alvarez, J.~M., Fidler, S., and Law, M.~T.
\newblock Optimizing data collection for machine learning.
\newblock \emph{arXiv preprint arXiv:2210.01234}, 2022.

\bibitem[NVIDIA(2021)]{nvdrivesim}
NVIDIA.
\newblock Nvidia drivesim, 2021.
\newblock URL \url{https://developer.nvidia.com/drive/simulation}.

\bibitem[Prabhu et~al.(2021)Prabhu, Khare, Kartik, and
  Hoffman]{prabhu2021sentry}
Prabhu, V., Khare, S., Kartik, D., and Hoffman, J.
\newblock Sentry: Selective entropy optimization via committee consistency for
  unsupervised domain adaptation.
\newblock In \emph{Proceedings of the IEEE/CVF International Conference on
  Computer Vision}, pp.\  8558--8567, 2021.

\bibitem[Prakash et~al.(2019)Prakash, Boochoon, Brophy, Acuna, Cameracci,
  State, Shapira, and Birchfield]{prakash2019structured}
Prakash, A., Boochoon, S., Brophy, M., Acuna, D., Cameracci, E., State, G.,
  Shapira, O., and Birchfield, S.
\newblock Structured domain randomization: Bridging the reality gap by
  context-aware synthetic data.
\newblock In \emph{2019 International Conference on Robotics and Automation
  (ICRA)}, pp.\  7249--7255. IEEE, 2019.

\bibitem[Prakash et~al.(2021)Prakash, Debnath, Lafleche, Cameracci, Birchfield,
  Law, et~al.]{prakash2021self}
Prakash, A., Debnath, S., Lafleche, J.-F., Cameracci, E., Birchfield, S., Law,
  M.~T., et~al.
\newblock Self-supervised real-to-sim scene generation.
\newblock In \emph{Proceedings of the IEEE/CVF International Conference on
  Computer Vision}, pp.\  16044--16054, 2021.

\bibitem[Ramamonjison et~al.(2021)Ramamonjison, Banitalebi-Dehkordi, Kang, Bai,
  and Zhang]{ramamonjison2021simrod}
Ramamonjison, R., Banitalebi-Dehkordi, A., Kang, X., Bai, X., and Zhang, Y.
\newblock Simrod: A simple adaptation method for robust object detection.
\newblock In \emph{Proceedings of the IEEE/CVF International Conference on
  Computer Vision}, pp.\  3570--3579, 2021.

\bibitem[Rempe et~al.(2022)Rempe, Philion, Guibas, Fidler, and
  Litany]{rempe2022generating}
Rempe, D., Philion, J., Guibas, L.~J., Fidler, S., and Litany, O.
\newblock Generating useful accident-prone driving scenarios via a learned
  traffic prior.
\newblock In \emph{Proceedings of the IEEE/CVF Conference on Computer Vision
  and Pattern Recognition}, pp.\  17305--17315, 2022.

\bibitem[Ren et~al.(2015)Ren, He, Girshick, and Sun]{ren2015faster}
Ren, S., He, K., Girshick, R., and Sun, J.
\newblock Faster r-cnn: Towards real-time object detection with region proposal
  networks.
\newblock \emph{Advances in neural information processing systems}, 28, 2015.

\bibitem[Resnick et~al.(2022)Resnick, Litany, Kar, Kreis, Lucas, Cho, and
  Fidler]{resnick2022causal}
Resnick, C., Litany, O., Kar, A., Kreis, K., Lucas, J., Cho, K., and Fidler, S.
\newblock Causal scene bert: Improving object detection by searching for
  challenging groups of data.
\newblock \emph{arXiv preprint arXiv:2202.03651}, 2022.

\bibitem[RoyChowdhury et~al.(2019)RoyChowdhury, Chakrabarty, Singh, Jin, Jiang,
  Cao, and Learned-Miller]{roychowdhury2019automatic}
RoyChowdhury, A., Chakrabarty, P., Singh, A., Jin, S., Jiang, H., Cao, L., and
  Learned-Miller, E.
\newblock Automatic adaptation of object detectors to new domains using
  self-training.
\newblock In \emph{Proceedings of the IEEE/CVF Conference on Computer Vision
  and Pattern Recognition}, pp.\  780--790, 2019.

\bibitem[Saenko et~al.(2010)Saenko, Kulis, Fritz, and
  Darrell]{saenko2010adapting}
Saenko, K., Kulis, B., Fritz, M., and Darrell, T.
\newblock Adapting visual category models to new domains.
\newblock In \emph{European conference on computer vision}, pp.\  213--226.
  Springer, 2010.

\bibitem[Saito et~al.(2019{\natexlab{a}})Saito, Kim, Sclaroff, Darrell, and
  Saenko]{saito2019semi}
Saito, K., Kim, D., Sclaroff, S., Darrell, T., and Saenko, K.
\newblock Semi-supervised domain adaptation via minimax entropy.
\newblock In \emph{Proceedings of the IEEE International Conference on Computer
  Vision}, pp.\  8050--8058, 2019{\natexlab{a}}.

\bibitem[Saito et~al.(2019{\natexlab{b}})Saito, Ushiku, Harada, and
  Saenko]{saito2019strong}
Saito, K., Ushiku, Y., Harada, T., and Saenko, K.
\newblock Strong-weak distribution alignment for adaptive object detection.
\newblock In \emph{Proceedings of the IEEE/CVF Conference on Computer Vision
  and Pattern Recognition}, pp.\  6956--6965, 2019{\natexlab{b}}.

\bibitem[Scott(2015)]{scott2015multivariate}
Scott, D.~W.
\newblock \emph{Multivariate density estimation: theory, practice, and
  visualization}.
\newblock John Wiley \& Sons, 2015.

\bibitem[Tremblay et~al.(2018)Tremblay, Prakash, Acuna, Brophy, Jampani, Anil,
  To, Cameracci, Boochoon, and Birchfield]{tremblay2018training}
Tremblay, J., Prakash, A., Acuna, D., Brophy, M., Jampani, V., Anil, C., To,
  T., Cameracci, E., Boochoon, S., and Birchfield, S.
\newblock Training deep networks with synthetic data: Bridging the reality gap
  by domain randomization.
\newblock In \emph{Proceedings of the IEEE conference on computer vision and
  pattern recognition workshops}, pp.\  969--977, 2018.

\bibitem[Tsai et~al.(2016)Tsai, Yeh, and Wang]{tsai2016learning}
Tsai, Y.-H.~H., Yeh, Y.-R., and Wang, Y.-C.~F.
\newblock Learning cross-domain landmarks for heterogeneous domain adaptation.
\newblock In \emph{Proceedings of the IEEE conference on computer vision and
  pattern recognition}, pp.\  5081--5090, 2016.

\bibitem[Tzeng et~al.(2017)Tzeng, Hoffman, Saenko, and
  Darrell]{tzeng2017adversarial}
Tzeng, E., Hoffman, J., Saenko, K., and Darrell, T.
\newblock Adversarial discriminative domain adaptation.
\newblock In \emph{Proceedings of the IEEE Conference on Computer Vision and
  Pattern Recognition}, pp.\  7167--7176, 2017.

\bibitem[Wang et~al.(2019)Wang, Zhang, Yuan, and Feng]{wang2019few}
Wang, T., Zhang, X., Yuan, L., and Feng, J.
\newblock Few-shot adaptive faster r-cnn.
\newblock In \emph{Proceedings of the IEEE/CVF Conference on Computer Vision
  and Pattern Recognition}, pp.\  7173--7182, 2019.

\bibitem[Wang et~al.(2020)Wang, Huang, Gonzalez, Darrell, and
  Yu]{wang2020frustratingly}
Wang, X., Huang, T., Gonzalez, J., Darrell, T., and Yu, F.
\newblock Frustratingly simple few-shot object detection.
\newblock In \emph{International Conference on Machine Learning}, pp.\
  9919--9928. PMLR, 2020.

\bibitem[Wang et~al.(2021)Wang, Huang, Liu, Yu, Wang, Gonzalez, and
  Darrell]{wang2021robust}
Wang, X., Huang, T.~E., Liu, B., Yu, F., Wang, X., Gonzalez, J.~E., and
  Darrell, T.
\newblock Robust object detection via instance-level temporal cycle confusion.
\newblock In \emph{Proceedings of the IEEE/CVF International Conference on
  Computer Vision}, pp.\  9143--9152, 2021.

\bibitem[Wrenninge \& Unger(2018)Wrenninge and Unger]{wrenninge2018synscapes}
Wrenninge, M. and Unger, J.
\newblock Synscapes: A photorealistic synthetic dataset for street scene
  parsing.
\newblock \emph{arXiv preprint arXiv:1810.08705}, 2018.

\bibitem[Yao et~al.(2015)Yao, Pan, Ngo, Li, and Mei]{yao2015semi}
Yao, T., Pan, Y., Ngo, C.-W., Li, H., and Mei, T.
\newblock Semi-supervised domain adaptation with subspace learning for visual
  recognition.
\newblock In \emph{CVPR}, 2015.

\bibitem[Zhong et~al.(2022)Zhong, Wang, Feng, Zhang, Sun, and
  Yokota]{zhong2022pica}
Zhong, C., Wang, J., Feng, C., Zhang, Y., Sun, J., and Yokota, Y.
\newblock Pica: Point-wise instance and centroid alignment based few-shot
  domain adaptive object detection with loose annotations.
\newblock In \emph{Proceedings of the IEEE/CVF Winter Conference on
  Applications of Computer Vision}, pp.\  2329--2338, 2022.

\bibitem[Zhu et~al.(2019)Zhu, Pang, Yang, Shi, and Lin]{zhu2019adapting}
Zhu, X., Pang, J., Yang, C., Shi, J., and Lin, D.
\newblock Adapting object detectors via selective cross-domain alignment.
\newblock In \emph{Proceedings of the IEEE/CVF Conference on Computer Vision
  and Pattern Recognition}, pp.\  687--696, 2019.

\end{thebibliography}
